\documentclass[10pt,twocolumn,letterpaper]{article}
\usepackage{fullpage}
\usepackage[dvipsnames]{xcolor}
\usepackage{hyperref}
\hypersetup{
    colorlinks=true,
    linkcolor=blue,
    citecolor=violet,
    filecolor=magenta,
    urlcolor=cyan,
}

\usepackage{authblk}
\usepackage{graphicx}
\usepackage{graphbox}
\graphicspath{{./figures/}}
\usepackage{wrapfig}
\usepackage{subcaption}

\usepackage{tikz-cd}

\usepackage{amssymb,amsmath,amsthm,bbm}
\usepackage{mathtools}%
\usepackage[capitalize]{cleveref}

\usepackage{amsthm,amssymb,amsmath}
\usepackage{cutwin}
\usepackage{enumitem}
\usepackage{microtype}

\numberwithin{equation}{section}

\Crefname{defn}{Defn.}{Defns.}
\Crefname{lemma}{Lem.}{Lems.}
\Crefname{alg}{Alg.}{Algs.}

\usepackage{thmtools}
\usepackage{thm-restate}

\usepackage[backend = biber, url=false,sorting=nyt,style=numeric-comp,maxbibnames = 99]{biblatex}
\usepackage[colorinlistoftodos,prependcaption,textsize=small]{todonotes}

\definecolor{ltgray}{gray}{0.9}
\definecolor{gray}{rgb}{0.95,0.95,0.96}
\definecolor{dkgray}{rgb}{0.7,0.7, 0.735}
\definecolor{ltblue}{rgb}{0.55,0.55, 0.95}
\definecolor{ltGreen}{rgb}{0.25,0.65, 0.25}
\definecolor{dkgreen}{RGB}{0, 100, 0}
\definecolor{dkred}{rgb}{0.75,0.0, 0.0}
\definecolor{dkorange}{rgb}{.82,.45, 0.0}
\definecolor{ltred}{rgb}{0.95,0.95, 0.85}
\definecolor{utahRed}{rgb}{.8, 0, 0}
\definecolor{oregonGreen}{rgb}{0, .41, .163}
\definecolor{albanyPurple}{rgb}{0.4, 0, .55}

\newcommand{\E}{\mathbb{E}}
\newcommand{\N}{\mathbb{N}}
\newcommand{\R}{\mathbb{R}}
\newcommand{\W}{\mathbb{W}}
\newcommand{\V}{\mathbb{V}}
\newcommand{\X}{{\mathbb X}}
\newcommand{\Y}{{\mathbb Y}}
\newcommand{\Z}{{\mathbb Z}}

\newcommand{\e}{\varepsilon}
\renewcommand{\phi}{\varphi}

\usepackage{authblk}

\begin{document}
\title{Automatic Tree Ring Detection using Jacobi Sets}

\author[1]{Kayla Makela}
\author[1,2]{Tim Ophelders} 
\author[3]{Michelle Quigley}
\author[1,4]{Elizabeth Munch%
\thanks{muncheli@msu.edu\\
The work of EM was funded in part by NSF-CCF-1907591.  The work of EM and DC was funded in part by NSF-DEB-1904267.}
}
\author[3,1]{Daniel Chitwood}
\author[5]{Asia Dowtin}

\affil[1]{Dept of Computational Mathematics, Science and Engineering, Michigan State University}
\affil[2]{Dept of Mathematics and Computer Science, TU Eindhoven}
\affil[3]{Dept of Horticulture, Michigan State University}
\affil[4]{Dept of Mathematics, Michigan State University}
\affil[5]{Dept of Forestry, Michigan State University}

\date{}

\maketitle
\begin{abstract}
    Tree ring widths are an important source of climatic and historical data, but measuring these widths typically requires extensive manual work.
    Computer vision techniques provide promising directions towards the automation of tree ring detection, but most automated methods still require a substantial amount of user interaction to obtain high accuracy.
    We perform analysis on 3D X-ray CT images of a cross-section of a tree trunk, known as a tree disk. 
    We present novel automated methods for locating the pith (center) of a tree disk, and ring boundaries.
    Our methods use a combination of standard image processing techniques and tools from topological data analysis.
    We evaluate the efficacy of our method for two different CT scans by comparing its results to manually located rings and centers and show that it is better than current automatic methods in terms of correctly counting each ring and its location.
    Our methods have several parameters, which we optimize experimentally by minimizing edit distances to the manually obtained locations.
\end{abstract}

\section{Introduction}
Tree ring widths provide important data sources for dendrochronologists, climatologists, archaeologists, and more. The patterns of widths between rings can be used to date the trees because they will line up between similar trees due to being subjected to the same climactic patterns~\cite{Ferguson1970}. This process of comparing width sequences is known as crossdating and allows dendrochronologists to assign a specific year to each annual growth ring. Further, the ring can give information about that year, such as the amount of precipitation or an environmental event such as a forest fire~\cite{Schweingruber2008}.  

However, obtaining tree ring measurements takes extensive manual work. There has been considerable progress towards automation of this task in the realms of computer vision and machine learning, but most still require some amount of user interaction, such as marking the center or a measurement path. Deformations in rings, doubled or missing rings, and cuts in the wood can make reliable recognition difficult~\cite{Biondi2020}. Further, features indicating ring boundaries may vary between types of trees. For example, trees near the equator where season climactic variations are minimal do not tend to produce annual rings~\cite{Speer2010}.

There are a variety of programs available which offer the measurement of tree rings from images. Some of the most prominent commercial software include LIGNOVISION\texttrademark~\cite{lignovision} and WinDENDRO\texttrademark~\cite{windendro}. However, these programs are closed-source and require substantial user interaction. There have been many open-source attempts based on varying methods of computer vision and image processing~\cite{Bunn2008,Lara2015,Shi2019}, deep learning~\cite{Fabijanska2018}, and GIS~\cite{Arenas2015}. However, these methods still struggle to produce highly accurate results without user interaction, usually requiring a manual marking of the center or a measurement path. 

\begin{figure}
    \begin{subfigure}{0.5\columnwidth}
    \centering
        \includegraphics[height = 1in]{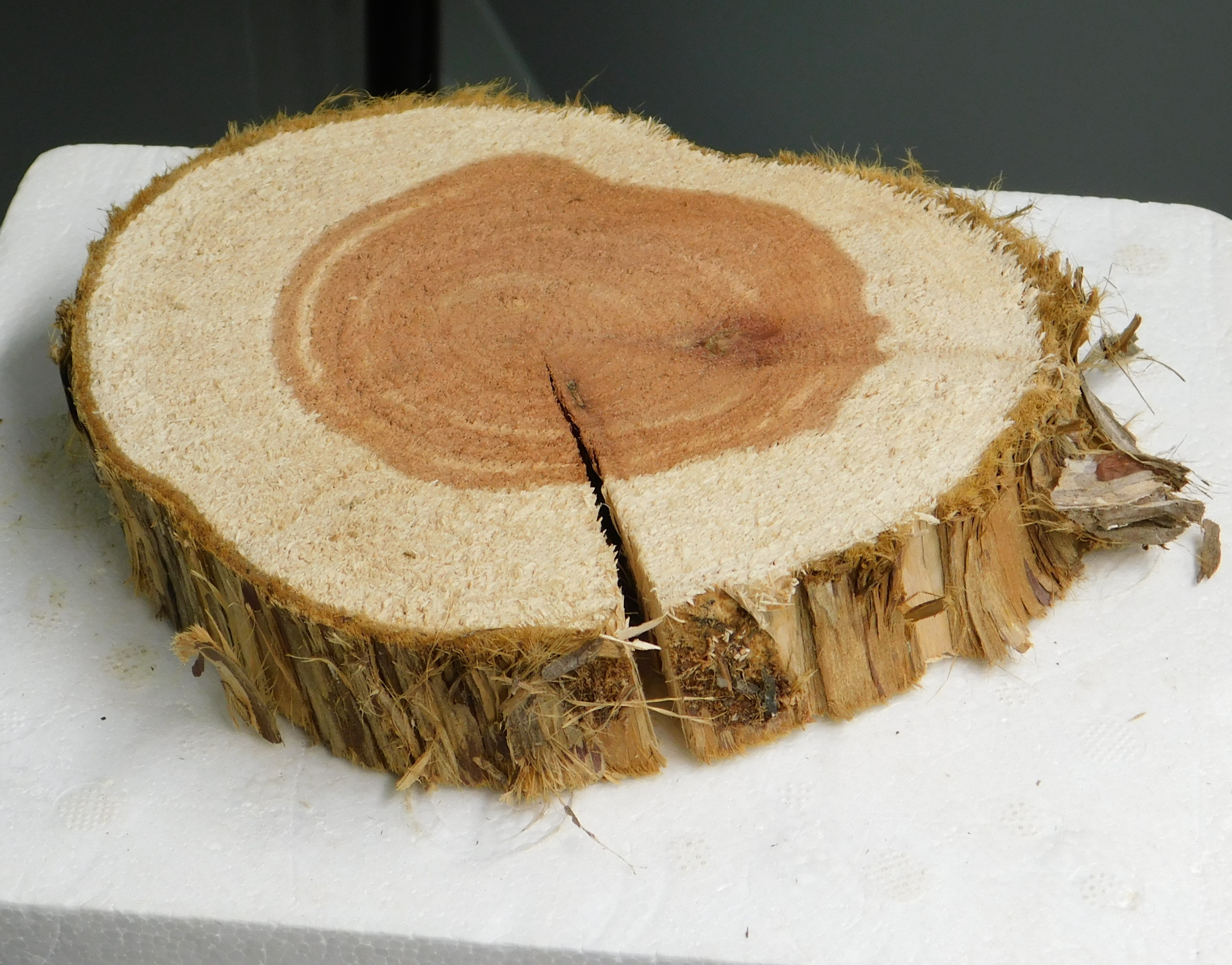}
        \includegraphics[height = 1.5in]{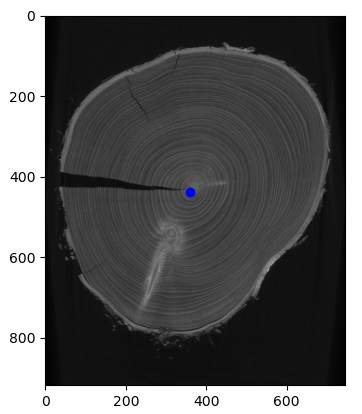}
        \caption{Red cedar disk.}
    \end{subfigure}%
    \begin{subfigure}{0.5\columnwidth}
    \centering
        \includegraphics[height = 1in]{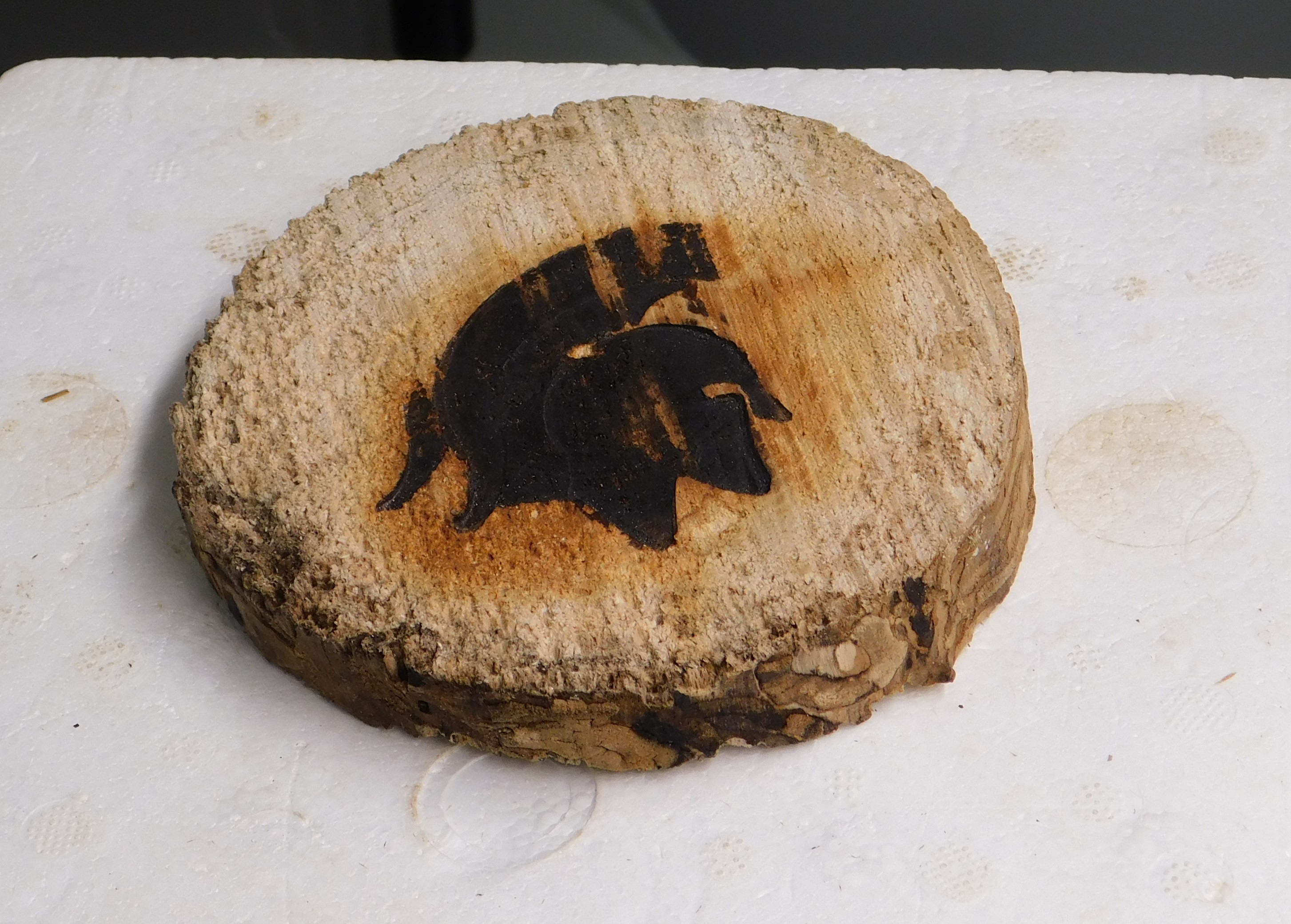}
        \includegraphics[height = 1.5in]{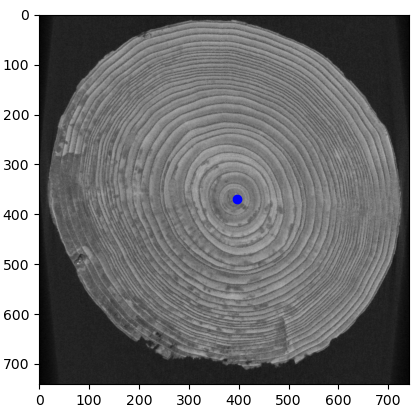}
        \caption{Spartan branded disk.}
    \end{subfigure}
    \caption{Images of the tree disk samples used to test the method with a slice of the 3D XRay CT scan shown in the bottom row. The blue dot shows the location of the center as determined by the method described in Sec.~\ref{sec:centerLocation}. }
    \label{fig:real}
\end{figure}

To remedy these issues, this paper presents an automated procedure for the full pipeline of tree ring analysis drawing from tools in topological data analysis \cite{Munch2017}. 
First, we utilize Sobel filters \cite{Kanopoulos1988} and region blurring for pith location, then we compute Jacobi sets for edge detection \cite{Edelsbrunner2004a}, and finally use persistent homology to refine the edge results \cite{Edelsbrunner2002}.
After pith detection, the located centers are refined using a line of best fit to ensure continuity between slices. 
2D slices of the 3D CT scan are converted to polar coordinates using their calculated centers to straighten out the rings for simpler edge recognition. Blurring and thresholding are used to clean up noise, each associated with a parameter which can be tuned to improve accuracy. 

The data used in this paper to test the algorithm consists of two 3D CT scans of different tree disks, labeled Red cedar and Spartan. 
These tree cores were obtained on the campus of Michigan State University and are shown in Figure~\ref{fig:real}.
The Red Cedar disk is labeled by its species; the species of the Spartan disk is undetermined, but it is labeled for its external branding which does not appear in the scans. 
We compare the results of our ring finding algorithm with the recent open-source program MtreeRing.

\section{Methodology}
Our method consists of two stages.
First, we locate the center of a 2D cross-section of a disk, using the algorithm presented in Section~\ref{sec:centerLocation}.
This center defines a radius for every point on the disk, and consequently defines the width of a tree ring at a particular angle.
The goal of the second algorithm (presented in Section~\ref{sec:ringLocation}) is to output a list of numbers indicating the position or width of rings on a given ray emanating from the center of the slice.

\subsection{Center Location}\label{sec:centerLocation}
\begin{figure}
    \centering
    \includegraphics[width=\columnwidth]{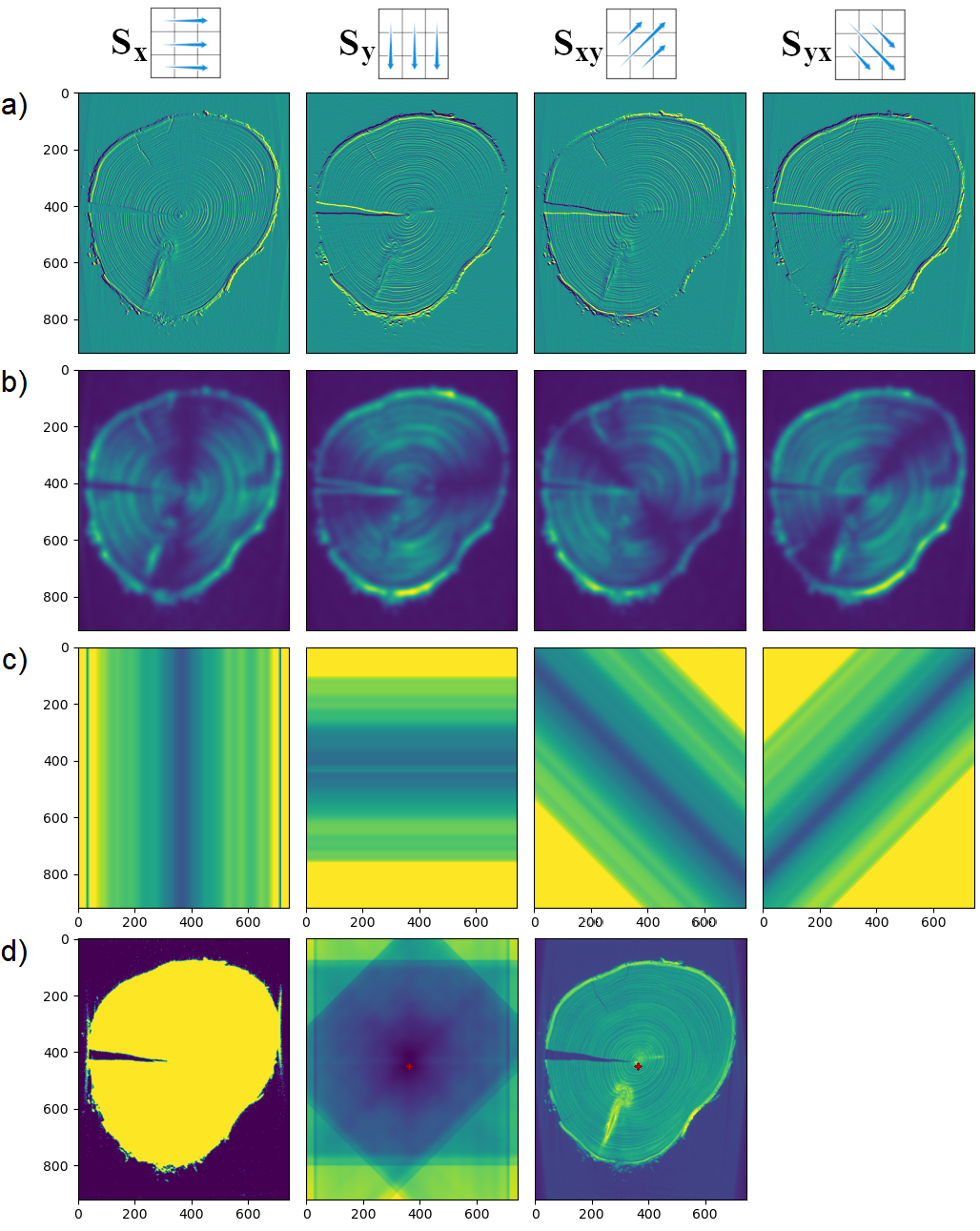}
    \caption{The center finding procedure. The top row shows the directions used for the Sobel filter, whose output is shown in row (a). 
    Row (b): blurred absolute values of row (a). 
    Row (c): average intensity of row (b) at corresponding coordinates.
    Row (d), left: mask used for averaging in row (c).
    Center: sum of row (c), with location of minimum marked.
    Right: location of minimum overlaid on original slice.
    }
    \label{fig:center}
\end{figure}
The first task is to find the center, or pith, of the given tree disk; an example of this procedure is shown in Fig.~\ref{fig:center}. 
Our proposed method of center location relies on Sobel filters, which is a derivative mask commonly used for feature detection in computer vision~\cite{Kanopoulos1988}. 
The goal of the Sobel filter is to approximate gradients in an image, so pixels with the highest magnitude of gradient can be designated as belonging to edges. 
The Sobel operator is defined in the horizontal and vertical direction, as well as the two diagonal directions by $3\times 3$ convolution kernels $S_x$, $S_y$, $S_{xy}$, and $S_{yx}$:
\begin{align*}
S_x &= 
\begin{pmatrix*}[r]
-1 & \hphantom{-}0 & \hphantom{-}1\\
-2 & 0 & 2\\
-1 & 0 & 1
\end{pmatrix*}\!\!,\!\!\!
&
S_{xy} = \frac{1}{\sqrt{2}}
\begin{pmatrix*}[r]
-2 & -2 & \hphantom{-}0\\
-2 & 0 & 2\\
0 & 2 & 2
\end{pmatrix*}\!\!,
\\
S_y &= 
\begin{pmatrix*}[r]
1 & 2 & 1\\
0 & 0 & 0\\
-1 & -2 & -1
\end{pmatrix*}\!\!,\!\!\!
&
S_{yx} = \frac{1}{\sqrt{2}}
\begin{pmatrix*}[r]
0 & 2 & \hphantom{-}2\\
-2 & 0 & 2\\
-2 & -2 & 0
\end{pmatrix*}\!\!.
\end{align*}

For a single two-dimensional slice, consider the Sobel filter based on $S_x$, shown in the top left of Fig.~\ref{fig:center}.
This filter assigns higher absolute values to vertical edges.
In the context of tree rings, we expect very few vertical edges at the $x$-coordinate of the center (since tree rings cross that $x$-coordinate in a perpendicular fashion), and more or longer vertical edges at $x$-coordinates further away from the tree center.
In particular, we expect the Sobel filter in a given direction to produce, on average, higher absolute values at coordinates (in that direction) away from the center; see row (b) of Fig.~\ref{fig:center}.
In order to ignore edges that occur in the background of the image, we identify the pixels that are part of the tree disk using a mask.
Such a mask can viably be obtained using a threshold intensity when using X-ray CT scans, as well as when working with pictures taken against a contrasting background.
The used mask is shown in the left column of row (d) of Fig.~\ref{fig:center}.

After blurring the absolute values resulting from the Sobel filter, we take, at a given $x$-coordinate, the average over all pixels that lie on the tree disk, see row (c) of Fig.~\ref{fig:center}.
If for a given $x$-coordinate, fewer than a threshold number of pixels (100 in our case) lie on the tree disk, we default to the maximum value over the image, rather than the average at that $x$-coordinate.
This results in a function that assigns to any $x$-coordinate, a quantity representing how vertically oriented the rings at that $x$-coordinate are.
We similarly compute a function that represents how horizontally oriented the rings are at a given $y$-coordinate, and similar functions for the two diagonal directions; shown in rows (a-c) of columns 2-4.

Intuitively, two such functions can be combined to locate the center of the disk: look for the $(x,y)$-coordinate with a minimum number of rings oriented vertically at the $x$-coordinate, and a minimum number of rings oriented horizontally at the $y$-coordinate.
However, considering only the $x$- and $y$- directions is not very robust, as radial cracks may form in disks cut from felled trees, as can be seen in the Red Cedar disk, Fig.~\ref{fig:real}.
This creates an edge that is oriented in the direction perpendicular to that of the rings at a given $x$- or $y$-coordinate, and may give the function an undesirably high value at the center of the disk.
To remedy this, we also use the diagonal directions, as it is unlikely for a disk to have cracks in multiple directions.
In our final step (see Fig.~\ref{fig:center}, row (d), columns 2 and 3), we sum for each pixel, the four directional functions at the corresponding coordinate, and identify the location of the pixel with minimum value as the center of the disk.

\begin{figure}
    \begin{subfigure}{0.4\linewidth}
        \includegraphics[scale=0.37]{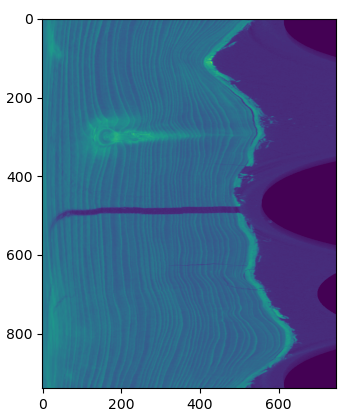}
        \caption{Red cedar.}
    \end{subfigure}%
    \begin{subfigure}{0.6\linewidth}
    \centering
        \includegraphics[scale=0.37]{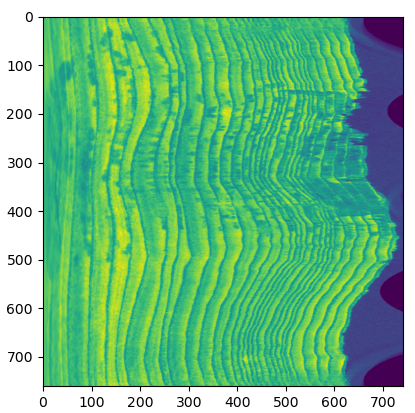}
        \caption{Spartan.}
    \end{subfigure}
    \caption{Padded polar images.}
    \label{fig:polars}
\end{figure}

In the case of a 3D image, we can repeat this process for every 2D slice, and take a line of best fit to ensure that the centers move continuously between adjacent slices.
Given the center of a particular 2D slice, we transform the slice into polar coordinates, so that $y$-coordinates correspond to angles, and $x$-coordinates correspond to distance from the center.
To avoid loss of information around the edges of the images, we pad the top of the image with rows from the bottom of the image.
These padded polar images represent the end of the preprocessing stage and are shown in Figure~\ref{fig:polars}.
\subsection{Ring Detection}\label{sec:ringLocation}
The next step in the tree ring analysis is to detect the ring boundaries, which is a problem of edge detection in image processing.

For our tree disk scans, we consider two functions associated with the polar image.
\begin{enumerate}
    \item The intensity value $f(x,y)$ of a 2D grayscale image can be interpreted as a continuous function (by interpolating between adjacent pixels) from a 2D domain.
    \item The $y$-coordinate $g(x,y):=y$ of pixels of a 2D image can similarly be interpreted as a continuous function with the same domain.
\end{enumerate}

\begin{figure}
        \includegraphics[scale=0.2,align = c]{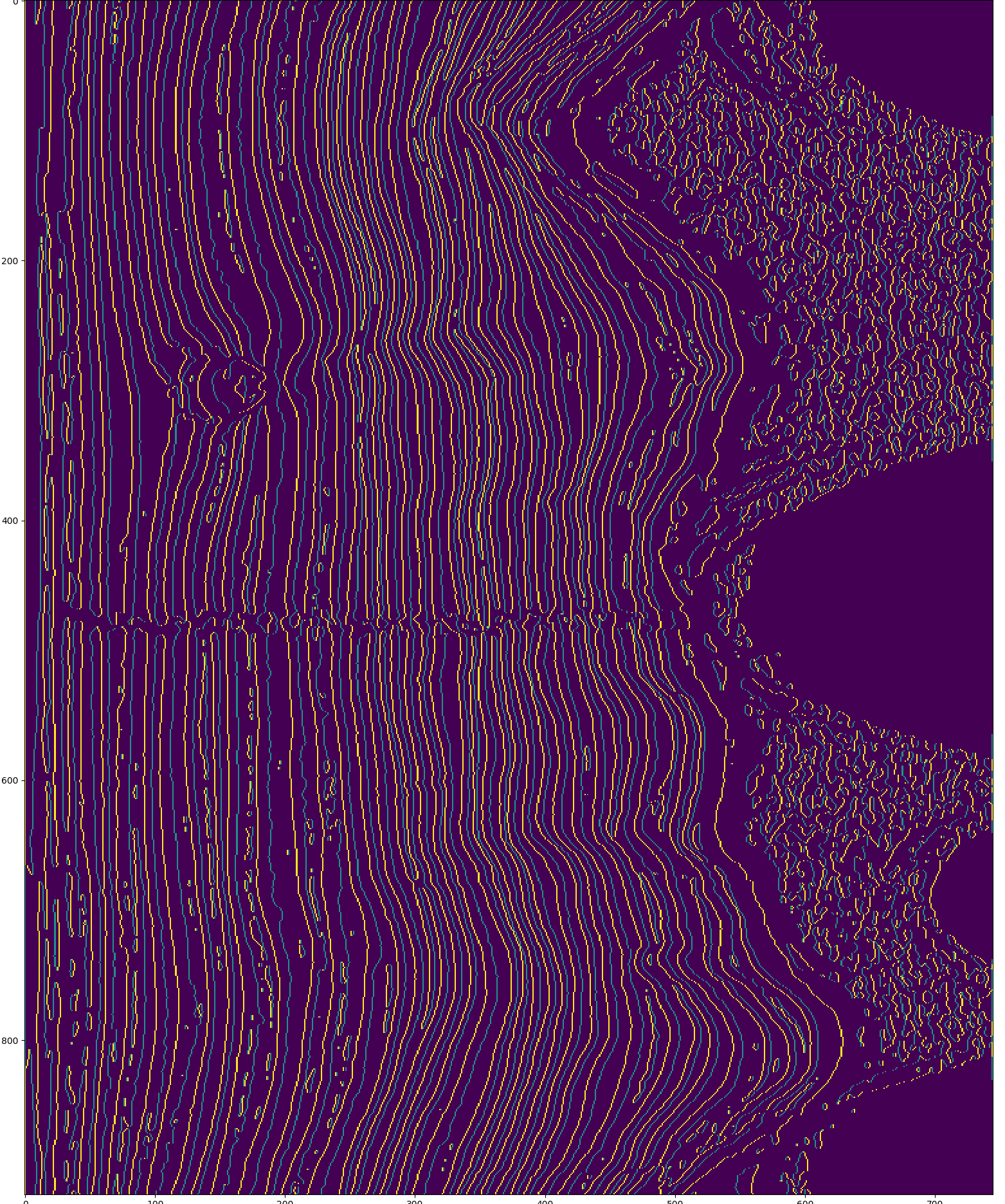}
        \includegraphics[scale=0.5,align=c]{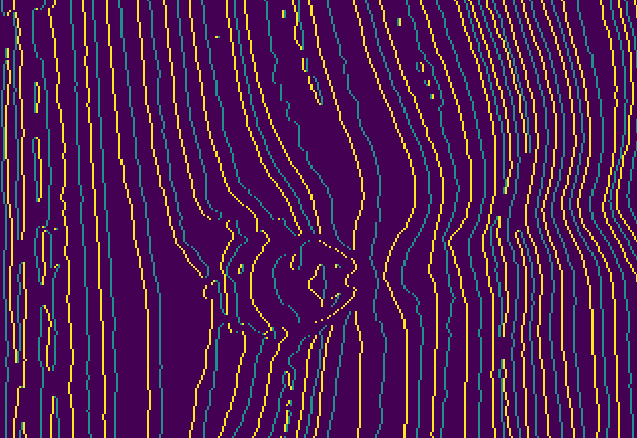}

    \caption{The results of the ring detection method. At top is the result of the Jacobi set detection method on Red cedar with a closeup shown below. Ridges in yellow corresponding to rings in this species; valleys in blue corresponding to rings in some species.
    }
    \label{fig:jacobi}
\end{figure}

To detect ring boundaries, we use a method based on Jacobi sets \cite{Edelsbrunner2004a}.
The Jacobi set of two real-valued functions with the same domain (such as the ones defined above) is defined as the set of critical points of one function restricted to the level sets of the other.
For our functions $f$ and $g$, the Jacobi set corresponds to (a) the local minima and maxima of intensity value at any fixed $y$-coordinate, as well as (b) the critical values of $g$ on components of a fixed intensity.
For two sufficiently well-behaved functions (i.e.~Morse functions), the Jacobi set forms a collection of paths \cite{Edelsbrunner2004a}, which we can think of as ridges and valleys of type (a), pairs of which end at points of type (b).
These ridges and valleys are shown for Red cedar in Fig.~\ref{fig:jacobi}.
In most conifer trees, the ridges correspond to ring boundaries.
However, in some species, such as the Spartan sample, the valleys indicate rings boundaries.

\begin{figure}
    \centering
    \includegraphics{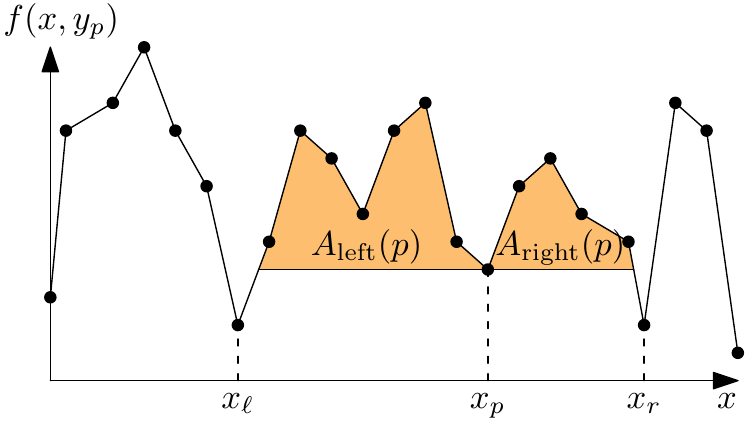}
    \caption{The areas that determine $p$'s area-based persistence.}
    \label{fig:areaPersistenceIllustration}
\end{figure}
Area-based persistence~\cite{Sonke} is applied to the Jacobi results as follows in order to clean up noise and assign high values to ridges based on their persistence.
Area-based persistence works by assigning a \emph{persistence value} to each pixel.
For a particular pixel $p$, this assignment is based on the intensities of pixels in its row (i.e. the restriction of $f$ to the $y$-coordinate of that pixel).
If the intensity of $p$ is not a local minimum (in its row), the persistence value of $p$ is zero.
On the other hand, if $p=(x_p,y_p)$ is a local minimum, we locate the closest pixels $\ell=(x_l,y_p)$ and $r=(x_r,y_p)$ to its left and right that have less intensity (so that $f(x,y_p)\geq f(x_p,y_p)$ for $x$ between $\ell$ and $r$), and compute the areas
\begin{align*}
    A_\text{left}(p)&=\int_{x_\ell}^{x_p}\max(0,f(x,y_p)-f(x_p,y_p))dx\text{, and}\\
    A_\text{right}(p)&=\int_{x_p}^{x_r}\max(0,f(x,y_p)-f(x_p,y_p))dx
\end{align*} under the intensity function above the intensity of $p$, see Fig.~\ref{fig:areaPersistenceIllustration}.
The persistence value of pixel $p$ will be the area of the smaller side: $\min\{A_\text{left}(p),A_\text{right}(p)\}$.
Intuitively, more significant local minima (valleys) will lie between mountains that are high (in terms of area), so significant minima will be assigned a high persistence value.
A symmetric method can be used to find persistent ridges (maxima).
The persistence value of a minimum (or maximum, depending on the type of tree) corresponds to the certainty with which we call it a tree ring boundary.
As such, we mark all minima or maxima that have at least some threshold persistence value as ring boundaries.

\begin{figure*}
    \centering
        \begin{subfigure}{0.49\textwidth}
        \centering
        \includegraphics[width = \textwidth]{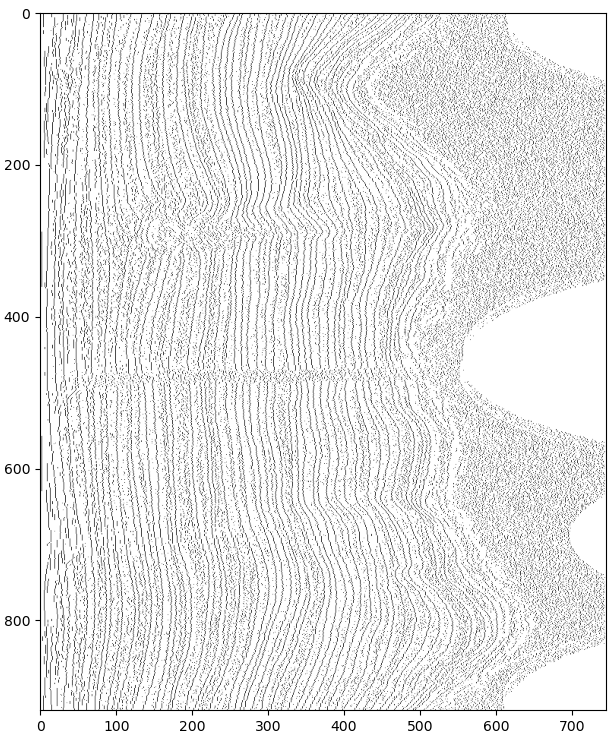}
        \caption{Blur 0, pre- and post- threshold 0.}
    \end{subfigure}
    \begin{subfigure}{0.49\textwidth}
        \centering
        \includegraphics[width = \textwidth]{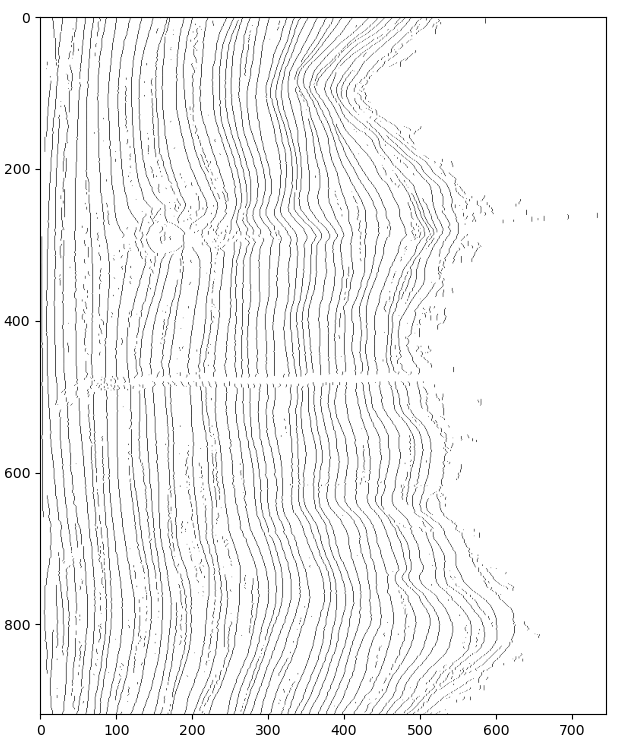}
        \caption{Blur 1, pre-threshold 0.12, post-threshold 0.02.}
    \end{subfigure}
    
    \begin{subfigure}{0.4\linewidth}
        \centering
        \includegraphics[width = \textwidth]{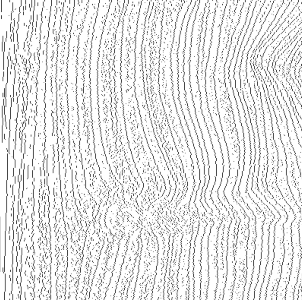}
        \caption{Closeup.}
    \end{subfigure}
    \begin{subfigure}{0.4\linewidth}
        \centering
        \includegraphics[width = \textwidth]{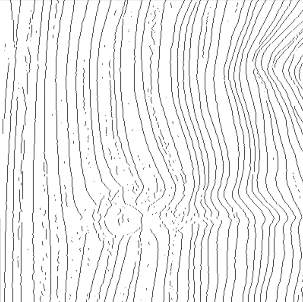}
        \caption{Closeup.}
    \end{subfigure}

    \caption{Red cedar results before and after persistence based filtering. At left are the detected local maxima with 0 for both blur and thresholding.  At right, the same slice with blur 1, pre-threshold 0.12, and post-threshold 0.02. 
    }
    \label{fig:pers_RedCedar}
\end{figure*}

\begin{figure}
    \begin{subfigure}{1.0\linewidth}
        \centering
        \includegraphics[scale=0.77]{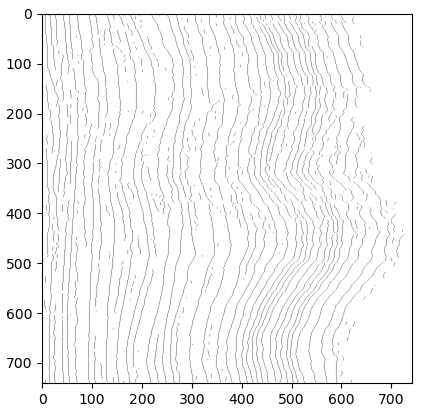}
        \caption{Full.}
    \end{subfigure}
    \begin{subfigure}{1.0\linewidth}
        \centering
        \includegraphics[scale=1.83]{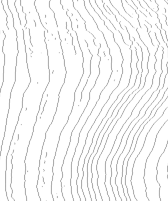}
        \caption{Closeup.}
    \end{subfigure}
    \caption{Spartan results with minimum cost. Blur 2, pre-threshold 0.16, post-threshold 0.}
    \label{fig:bestspartan}
\end{figure}

In the next section, we explore different threshold values, and validate our method against manually called ring boundaries, as well as state-of-the-art software.
Results of persistence with different choices of thresholds and blur parameters are shown in Fig.~\ref{fig:pers_RedCedar} for the red cedar disk, and in Fig.~\ref{fig:bestspartan} for the Spartan disk.

\section{Evaluation of Results}
The end goal of the algorithm is to output a list of numbers indicating the positions or widths of rings along a given path through a slice of the tree core. 
We test our results against the ground truth, created  by manually marking ring boundaries on the image using ImageJ~\cite{imagej} software. 
Thus, we next need to compare these two lists, being vigilant for falsely identified or missing rings. 
There are three parameters that impact the performance of the algorithm: the blur value during the persistence process and two thresholds, before and after edge detection. We develop a cost function based on string edit distances to compare various combinations of parameters. 

\subsection{Tree Ring Edit Distance}
We use a traditional form of the edit distance~\cite{Wenk1999} to compare sequences of tree ring widths. 
For each run of the scoring code, two text lists of x-values are compared using matching and costs based on distance. 
Once pairs are matched, the algorithm places a cost on each of three operations: 200 to add a point (resolve false negative), 200 to remove a point (resolve false positive), and the distance in pixels between two points to move a point (resolve minor errors in placement). 
The choice of 200 is high enough to make missing or false rings the dominant factor in the cost, and low enough to still be affected by placement errors.  
Note that the positions of the rings in the ground truth data set are marked by the researcher and thus are also subject to human error. 

\begin{figure}
    \centering 
    \begin{subfigure}{1.0\linewidth}
        \includegraphics[scale=0.58]{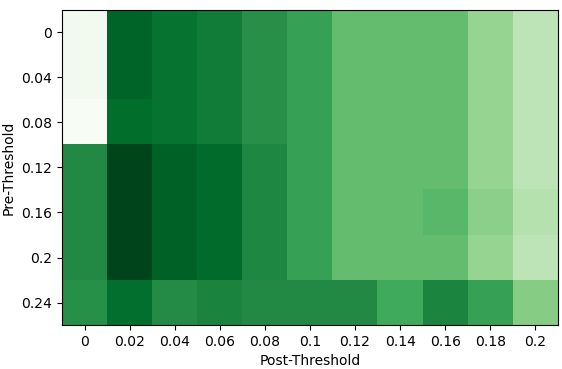}
        \caption{Red cedar threshold combinations with blur fixed at 1.}
    \end{subfigure}
    \begin{subfigure}{1.0\linewidth}
        \includegraphics[scale=0.58]{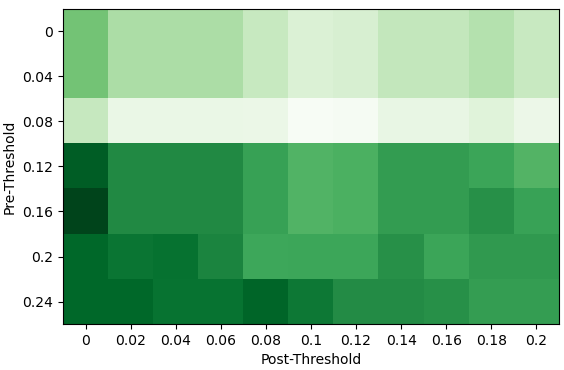}
        \caption{Spartan threshold combinations with blur fixed at 2.}
    \end{subfigure}
    \caption{The heat maps show the edit distance between the computed ring locations and the ground truth for different parameter combinations; darker is lower cost, thus meaning that choice of parameters does a better job of approximating the ground truth locations. Best performance for Red cedar comes with blur 1, pre-threshold 0.12, and post-threshold 0.02. Best performance for Spartan comes with blur 2, pre-threshold 0.16, and post-threshold 0.
    }
    \label{fig:heatmaps}
\end{figure}

\subsection{Parameter Choices}
\begin{figure}
    \begin{subfigure}{1.0\linewidth}
        \centering
        \includegraphics[scale=0.35]{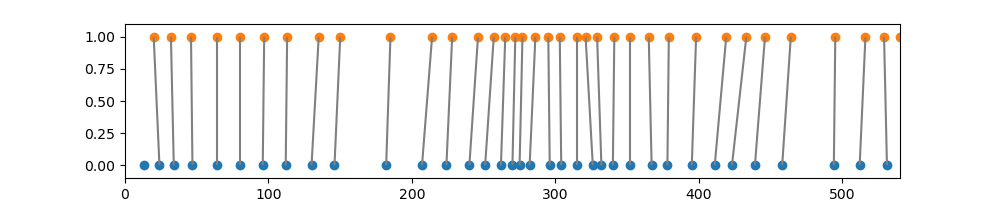}
        \caption{Red Cedar, our algorithm (cost 964).}
    \end{subfigure}
    \begin{subfigure}{1.0\linewidth}
        \centering
        \includegraphics[scale=0.35]{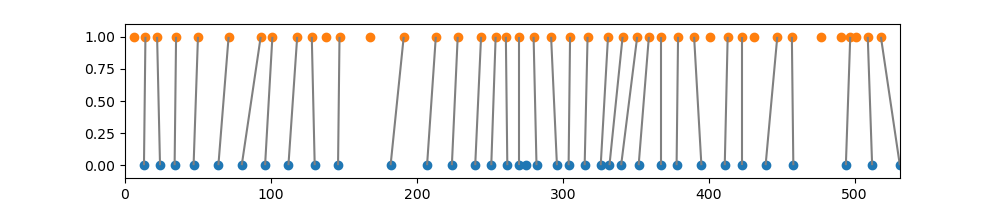}
        \caption{Red Cedar, MtreeRing LinearDetect (cost 2841).}
    \end{subfigure}
        \begin{subfigure}{1.0\linewidth}
        \centering
        \includegraphics[scale=0.35]{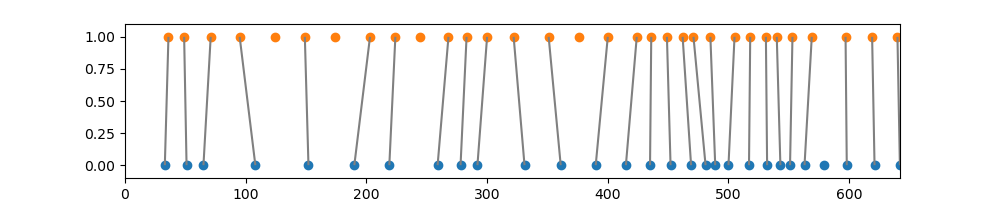}
        \caption{Spartan, our algorithm (cost 2213).}
    \end{subfigure}
    \begin{subfigure}{1.0\linewidth}
        \centering
        \includegraphics[scale=0.35]{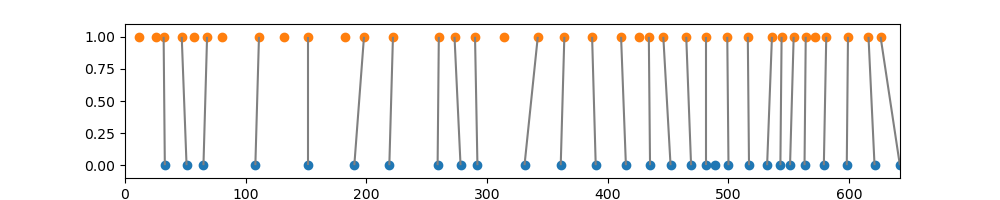}
        \caption{Spartan, MtreeRing LinearDetect (cost 2661).}
    \end{subfigure}
    \caption{Comparing the best results of our algorithm to the best results of MtreeRing for both disks. Matchings between the human-annotated ground truth (blue) and computed ring locations (orange).}
    \label{fig:bestmatching}
\end{figure}

We examine edit distance costs for various combinations of the three aforementioned parameters: pre-threshold, blur, and post-threshold. Both thresholds range from $0$ to $0.2\cdot n$ where~$n$ is the maximum intensity value in the image. The blur value is the value of $\sigma$ in a Gaussian blur filter, which indicates the standard deviation for the Gaussian kernel. Exploring these options between species allows us to consider which parameters can be fixed and which will need to be input or determined by the image. Costs for various parameter combinations with fixed blur value are shown in Figure~\ref{fig:heatmaps}. A higher blur value for the Spartan sample performs better, likely due to the increased space between rings as compared to the Red Cedar sample. The results with the minimum cost for Red Cedar are shown in Figure~\ref{fig:pers_RedCedar}, and for Spartan in Figure~\ref{fig:bestspartan}.
The matching for these results is shown in Figure~\ref{fig:bestmatching}.

\subsection{Comparison with MtreeRing}
MtreeRing is a newer open-source program in R for tree ring analysis which compares favorably with prominent software such as WinDENDRO\texttrademark~\cite{Shi2019}. We compare our results with MtreeRing using the edit distance cost we have developed. 

For the same row on the same slice of each tree sample, we compute the tree ring boundaries using both our method and MtreeRing. For the Red Cedar sample, the output of our program has a cost of 964 whereas the MtreeRing LinearDetect output costs 2841. For the Spartan sample, our cost is 2213 and MtreeRing has cost 2661.

\section{Conclusion}
This paper has presented a new  method of automatic center and ring detection for scans of tree cores. 
The most far reaching effects will come from the fact that very little human input is necessary, particularly comparing with the state of the art software, the most accurate of which requires a human-drawn line to use in computations.
We are currently working to incorporate improvements to this method can be made in terms of accuracy and applicability across tree species, specifically in terms of ring boundary detection. 
First, care will need to be taken to consider missing or double rings which will likely require significant collaboration with dendrochronologists. 
To avoid choosing a measurement path, which may run into knots in the wood or other obstacles, a study of volume between rings could be made in three dimensions. 
Nonetheless, these computational advancements in dendrochronology have the potential to speed up tree ring analysis and therefore provide more data to climatologists and others for further understanding of climate change.

\printbibliography

\end{document}